\documentclass[11pt,a4paper]{article}
\usepackage[hyperref]{acl2021}
\usepackage{times}
\usepackage{latexsym}

\usepackage{microtype}

\usepackage{multirow}
\usepackage{makecell}
\usepackage{url}
\usepackage{subcaption}
\usepackage[export]{adjustbox} 
\usepackage{graphicx}
\usepackage{amsmath}
\usepackage{booktabs}
\usepackage{pgf, tikz}
\usepackage{collcell}
\usetikzlibrary{arrows, automata}
\usepackage{xcolor}
\usepackage{xspace}

\newcommand{\sidecaption}[1]% #1 = label name
{{\begin{subfigure}[t]{1.6em}
  \caption[singlelinecheck=off]{}% do not center
\end{subfigure}}\ignorespaces}

\newcommand{\ba}{\textsc{b\&a}\xspace}
\colorlet{colorleft}{red!25}
\colorlet{colorright}{red!100}
\definecolor{color_c}{RGB}{211,211,211}
\definecolor{cyann}{RGB}{175, 247, 252}

\definecolor{mycolor0}{RGB}{237,104,60}
\definecolor{mycolor1}{RGB}{243,144,63}
\definecolor{mycolor2}{RGB}{253,199,12}
\definecolor{mycolor3}{RGB}{255,243,59}

\definecolor{tikzc1}{HTML}{207847}
\definecolor{tikzc2}{HTML}{2eac66}
 \definecolor{tikzc3}{HTML}{7ebc57}
 \definecolor{tikzc4}{HTML}{b3cd41}
 \definecolor{tikzc5}{HTML}{D1D931}
\definecolor{tikzc6}{HTML}{EFE521}

\aclfinalcopy % Uncomment this line for the final submission
\title{Shortformer: Better Language Modeling Using Shorter Inputs}

\author{Ofir Press$^{1,2}$  \quad Noah A. Smith$^{1,3}$ \quad Mike Lewis$^2$ \\
\\ $^1$Paul G. Allen School of Computer Science \& Engineering, University of Washington
\\ $^2$Facebook AI Research
\\ $^3$Allen Institute for AI
\\ \texttt{ofirp@cs.washington.edu}
}

\date{}

\begin{document}
\maketitle
\begin{abstract}
Increasing the input length has been a driver of progress in language modeling with transformers.  We identify conditions where shorter inputs are not harmful, and achieve perplexity and efficiency improvements through two new methods that \emph{decrease} input length.
First, we show that initially training a model on short subsequences before moving on to longer ones both reduces overall training time and, surprisingly, substantially improves perplexity.
Second, we show how to improve the efficiency of recurrence methods in transformers, which let models condition on previously processed tokens when generating sequences that exceed the maximal length the transformer can handle at once. Existing methods require computationally expensive relative position embeddings; we introduce a simple alternative of adding absolute position embeddings to queries and keys instead of to word embeddings, which efficiently produces superior results. We show that these recurrent models also benefit from short input lengths. 
Combining these techniques speeds up training by a factor of 1.65, reduces memory usage, and substantially improves perplexity on WikiText-103, without adding any parameters.\footnote{Our code is available at \url{https://github.com/ofirpress/shortformer}}

\end{abstract}

\section{Introduction} \label{sec:intro}
Scaling up transformer~\cite{aiayn} language models \cite{gpt2,bart,t5,gpt3} has been an important driver of progress in NLP. 
Language models require data to be segmented into subsequences for both training and inference:  memory constraints limit a language model to handling at most a few thousand tokens at once, while many training and evaluation datasets are much longer. 
Recent work focuses on increasing the length of input subsequences, which determines the maximum number of tokens a model can attend to \cite{baevski, sukhbaatar2019adaptive, kitaev2020reformer, roy2020efficient}.

We challenge the assumption that longer input subsequences are \emph{always} better by showing that existing transformers do not always effectively use them. 
We then introduce new methods based on shorter input subsequences that improve runtime, memory efficiency, and perplexity.

We first investigate how input subsequence length affects transformer language models (\S\ref{sec:instance-size-matters}). 
Na\"{i}ve evaluation---where we split a large evaluation set into multiple nonoverlapping subsequences, each evaluated independently---initially supports the commonly-held belief that models that train and do inference on longer subsequences achieve better perplexity (Table~\ref{tab:bl_contextwindow}, col. 3). 

However, when we evaluate each model with a sliding window~\citep{baevski}, outputting one token at a time using the maximal amount of context, we find---surprisingly---that models using subsequences exceeding $1{,}024$ tokens do not further improve performance (Table~\ref{tab:bl_contextwindow}, col. 5).

We conclude that the performance gains (using na\"{i}ve evaluation) of models that use longer subsequences occur not only because of their better modeling ability, but partly because they divide the evaluation  set into longer subsequences. This division helps because of an issue we call the \emph{early token curse}: by default, early tokens in a subsequence will have short histories to attend to. 
Using longer subsequences means fewer tokens will suffer from the early token curse. For example, when using inputs of length $1{,}024$, about 94\% of tokens get to attend to more than 64 preceding tokens. If we use inputs of length $128$, only 50\% of tokens get to attend to 64 or more preceding tokens.

Based on this analysis, we explore how to improve models by using shorter inputs. We introduce two techniques.

\paragraph{Staged Training (\S\ref{sec:ss})}

First, we show that initially training on shorter subsequences (before moving to longer ones) leads not only to much faster and more memory-efficient training, but it surprisingly also greatly improves perplexity, suggesting that  longer inputs are  harmful early in training.

\paragraph{Position-Infused Attention (\S\ref{sec:pia})} 
Second,  we consider a natural way to avoid the early token curse during training and inference:   attending to cached representations from the previously evaluated subsequence~\cite{transformer-xl}. 
This approach interferes with conventional absolute position embeddings in a way that forced Dai et al. to use \emph{relative} position embeddings, which are computationally expensive.
We introduce a fast, simple alternative: instead of adding absolute position embeddings to word embeddings---thereby entangling a word's content and positional information---we  add them to the keys and queries in the self-attention mechanism (but \emph{not} to the values). This does not increase parameter count or runtime. Token representations can then be cached and reused in subsequent computations. 
We show that when using this method, shorter subsequence models outperform longer ones.

Finally, we show additive gains from combining staged training and position-infused attention (Shortformer, \S\ref{sec:combined}), resulting in a model that trains much quicker and achieves better perplexity on WikiText-103. We also show that these results transfer to language modeling on the Toronto Book Corpus (\S\ref{sec:books}, appendix).

\begin{figure*}
  \centering

  \sidecaption{subfig:a}
  \raisebox{-\height}{\scalebox{0.42}{\begin{tikzpicture}[
            %> = stealth, % arrow head style
            %shorten > = 1pt, % don't touch arrow head to node
            auto,
            node distance = 1cm, % distance between nodes
            semithick, % line style
        ]

        \tikzstyle{state1}=[
            circle,
            draw = tikzc1,
            fill = tikzc1,
            minimum size = 1mm,
        ]
        
        \tikzstyle{state2}=[
            circle,
            draw = tikzc2,
            fill = tikzc2,
            minimum size = 1mm,
        ]
        
        \tikzstyle{state3}=[
            circle,
            draw = tikzc3,
            fill = tikzc3,
            minimum size = 1mm,
        ]
        \tikzstyle{state4}=[
            circle,
            draw = tikzc4,
            fill = tikzc4,
            minimum size = 1mm,
        ]
        \tikzstyle{state5}=[
            circle,
            draw =tikzc5,
            fill =tikzc5,
            minimum size = 1mm,
        ]
        \tikzstyle{state6}=[
            circle,
            draw = tikzc6,
            fill = tikzc6,
            minimum size = 1mm,
        ]
\node[state1] (l0ts0)               [              label={[label distance=0.05cm,text=black]below:{\huge $a_1$ \vphantom{\huge {f}$_3$}  }}] {};
\node[state2] (l0ts1)  [right of=l0ts0,            label={[label distance=0.05cm,text=black]below:{\huge $b_2$ \vphantom{\huge {f}$_3$}  }}] {};
\node[state3] (l0ts2)  [right of=l0ts1,            label={[label distance=0.05cm,text=black]below:{\huge $c_3$ \vphantom{\huge {f}$_3$}  }}] {};
\node[state4] (l0ts3)  [right of=l0ts2,xshift=1cm, label={[label distance=0.05cm,text=black]below:{\huge $d_1$ \vphantom{\huge {f}$_3$}  }}] {};
\node[state5] (l0ts4)  [right of=l0ts3,            label={[label distance=0.05cm,text=black]below:{\huge $e_2$ \vphantom{\huge {f}$_3$}  }}] {};
\node[state6] (l0ts5)  [right of=l0ts4,            label={[label distance=0.05cm,text=black]below:{\huge $f_3$ \vphantom{\huge {f}$_3$}  }}] {};

\node[state1] (l1ts0)  [above of=l0ts0] {};
\node[state2] (l1ts1)  [above of=l0ts1] {};
\node[state3] (l1ts2)  [above of=l0ts2] {};
\node[state4] (l1ts3)  [above of=l0ts3] {};
\node[state5] (l1ts4)  [above of=l0ts4] {};
\node[state6] (l1ts5)  [above of=l0ts5] {};

\path[color=gray!30] [-] (l0ts0) edge node {}  (l1ts0);
\path[color=gray!30] [-] (l0ts0) edge node {}  (l1ts1);
\path[color=gray!30] [-] (l0ts1) edge node {}  (l1ts1);
\path[color=gray!30] [-] (l0ts0) edge node {}  (l1ts2);
\path[color=gray!30] [-] (l0ts1) edge node {}  (l1ts2);
\path[color=gray!30] [-] (l0ts2) edge node {}  (l1ts2);
\path[color=gray!30] [-] (l0ts3) edge node {}  (l1ts3);
\path[color=gray!30] [-] (l0ts3) edge node {}  (l1ts4);
\path[color=gray!30] [-] (l0ts4) edge node {}  (l1ts4);
\path[color=gray!30] [-] (l0ts3) edge node {}  (l1ts5);
\path[color=gray!30] [-] (l0ts4) edge node {}  (l1ts5);
\path[color=gray!30] [-] (l0ts5) edge node {}  (l1ts5);

\end{tikzpicture}}}%
  \quad
  \sidecaption{subfig:b}
  \raisebox{-\height}{\scalebox{0.42}{\begin{tikzpicture}[
            %> = stealth, % arrow head style
            %shorten > = 1pt, % don't touch arrow head to node
            auto,
            node distance = 1cm, % distance between nodes
            semithick, % line style
        ]

        \tikzstyle{state1}=[
            circle,
            draw = tikzc1,
            fill = tikzc1,
            minimum size = 1mm,
        ]
        
        \tikzstyle{state2}=[
            circle,
            draw = tikzc2,
            fill = tikzc2,
            minimum size = 1mm,
        ]
        
        \tikzstyle{state3}=[
            circle,
            draw = tikzc3,
            fill = tikzc3,
            minimum size = 1mm,
        ]
        \tikzstyle{state4}=[
            circle,
            draw = tikzc4,
            fill = tikzc4,
            minimum size = 1mm,
        ]
        \tikzstyle{state5}=[
            circle,
            draw =tikzc5,
            fill =tikzc5,
            minimum size = 1mm,
        ]
        \tikzstyle{state6}=[
            circle,
            draw = tikzc6,
            fill = tikzc6,
            minimum size = 1mm,
        ]h
\node[state1] (l0ts0)               [              label={[label distance=0.05cm,text=black]below:{\huge $a_1$ \vphantom{\huge {f}$_3$}  }}] {};
\node[state2] (l0ts1)  [right of=l0ts0,            label={[label distance=0.05cm,text=black]below:{\huge $b_2$ \vphantom{\huge {f}$_3$}  }}] {};
\node[state3] (l0ts2)  [right of=l0ts1,            label={[label distance=0.05cm,text=black]below:{\huge $c_3$ \vphantom{\huge {f}$_3$}  }}] {};
\node[state2] (l0ts3)  [right of=l0ts2,xshift=1cm, label={[label distance=0.05cm,text=black]below:{\huge $b_1$ \vphantom{\huge {f}$_3$}  }}] {};
\node[state3] (l0ts4)  [right of=l0ts3,            label={[label distance=0.05cm,text=black]below:{\huge $c_2$ \vphantom{\huge {f}$_3$}  }}] {};
\node[state4] (l0ts5)  [right of=l0ts4,            label={[label distance=0.05cm,text=black]below:{\huge $d_3$ \vphantom{\huge {f}$_3$}  }}] {};
\node[state3] (l0ts6)  [right of=l0ts5,xshift=1cm, label={[label distance=0.05cm,text=black]below:{\huge $c_1$ \vphantom{\huge {f}$_3$}  }}] {};
\node[state4] (l0ts7)  [right of=l0ts6,            label={[label distance=0.05cm,text=black]below:{\huge $d_2$ \vphantom{\huge {f}$_3$}  }}] {};
\node[state5] (l0ts8)  [right of=l0ts7,            label={[label distance=0.05cm,text=black]below:{\huge $e_3$ \vphantom{\huge {f}$_3$}  }}] {};
\node[state4] (l0ts9)  [right of=l0ts8,xshift=1cm, label={[label distance=0.05cm,text=black]below:{\huge $d_1$ \vphantom{\huge {f}$_3$}  }}] {};
\node[state5] (l0ts10) [right of=l0ts9,            label={[label distance=0.05cm,text=black]below:{\huge $e_2$ \vphantom{\huge {f}$_3$}  }}] {};
\node[state6] (l0ts11)[right of=l0ts10,            label={[label distance=0.05cm,text=black]below:{\huge $f_3$ \vphantom{\huge {f}$_3$}  }}] {};
\node[state1] (l1ts0)  [above of=l0ts0] {};
\node[state2] (l1ts1)  [above of=l0ts1] {};
\node[state3] (l1ts2)  [above of=l0ts2] {};
\node[state2] (l1ts3)  [above of=l0ts3] {};
\node[state3] (l1ts4)  [above of=l0ts4] {};
\node[state4] (l1ts5)  [above of=l0ts5] {};
\node[state3] (l1ts6)  [above of=l0ts6] {};
\node[state4] (l1ts7)  [above of=l0ts7] {};
\node[state5] (l1ts8)  [above of=l0ts8] {};
\node[state4] (l1ts9)  [above of=l0ts9] {};
\node[state5] (l1ts10)[above of=l0ts10] {};
\node[state6] (l1ts11)[above of=l0ts11] {};
\path[color=gray!30] [-] (l0ts0) edge node {}  (l1ts0);
\path[color=gray!30] [-] (l0ts0) edge node {}  (l1ts1);
\path[color=gray!30] [-] (l0ts1) edge node {}  (l1ts1);
\path[color=gray!30] [-] (l0ts0) edge node {}  (l1ts2);
\path[color=gray!30] [-] (l0ts1) edge node {}  (l1ts2);
\path[color=gray!30] [-] (l0ts2) edge node {}  (l1ts2);
\path[color=gray!30] [-] (l0ts3) edge node {}  (l1ts3);
\path[color=gray!30] [-] (l0ts3) edge node {}  (l1ts4);
\path[color=gray!30] [-] (l0ts4) edge node {}  (l1ts4);
\path[color=gray!30] [-] (l0ts3) edge node {}  (l1ts5);
\path[color=gray!30] [-] (l0ts4) edge node {}  (l1ts5);
\path[color=gray!30] [-] (l0ts5) edge node {}  (l1ts5);
\path[color=gray!30] [-] (l0ts6) edge node {}  (l1ts6);
\path[color=gray!30] [-] (l0ts6) edge node {}  (l1ts7);
\path[color=gray!30] [-] (l0ts7) edge node {}  (l1ts7);
\path[color=gray!30] [-] (l0ts6) edge node {}  (l1ts8);
\path[color=gray!30] [-] (l0ts7) edge node {}  (l1ts8);
\path[color=gray!30] [-] (l0ts8) edge node {}  (l1ts8);
\path[color=gray!30] [-] (l0ts9) edge node {}  (l1ts9);
\path[color=gray!30] [-] (l0ts9) edge node {}  (l1ts10);
\path[color=gray!30] [-] (l0ts10) edge node{}  (l1ts10);
\path[color=gray!30] [-] (l0ts9) edge node {}  (l1ts11);
\path[color=gray!30] [-] (l0ts10) edge node{}  (l1ts11);
\path[color=gray!30] [-] (l0ts11) edge node{}  (l1ts11);

\end{tikzpicture}}}
  \quad
  \sidecaption{subfig:c}
  \raisebox{-\height}{\scalebox{0.42}{\begin{tikzpicture}[
            %> = stealth, % arrow head style
            %shorten > = 1pt, % don't touch arrow head to node
            auto,
            node distance = 1cm, % distance between nodes
            semithick, % line style
        ]

        \tikzstyle{state1}=[
            circle,
            draw = tikzc1,
            fill = tikzc1,
            minimum size = 1mm,
        ]
        
        \tikzstyle{state2}=[
            circle,
            draw = tikzc2,
            fill = tikzc2,
            minimum size = 1mm,
        ]
        
        \tikzstyle{state3}=[
            circle,
            draw = tikzc3,
            fill = tikzc3,
            minimum size = 1mm,
        ]
        \tikzstyle{state4}=[
            circle,
            draw = tikzc4,
            fill = tikzc4,
            minimum size = 1mm,
        ]
        \tikzstyle{state5}=[
            circle,
            draw =tikzc5,
            fill =tikzc5,
            minimum size = 1mm,
        ]
        \tikzstyle{state6}=[
            circle,
            draw = tikzc6,
            fill = tikzc6,
            minimum size = 1mm,
        ]

\node[state1] (l0ts0)               [                  label={[label distance=0.05cm,text=black]below:{\huge {$a_1$} \vphantom{\huge {f}$_3$}  }}] {};
\node[state2] (l0ts1)  [right of=l0ts0,            label={[label distance=0.05cm,text=black]below:{\huge $b_2$ \vphantom{\huge {f}$_3$}  }}] {};
\node[state3] (l0ts2)  [right of=l0ts1,            label={[label distance=0.05cm,text=black]below:{\huge $c_3$ \vphantom{\huge {f}$_3$}  }}] {};
\node[state4] (l0ts3)  [right of=l0ts2,xshift=1cm, label={[label distance=0.05cm,text=black]below:{\huge $d_4$ \vphantom{\huge {f}$_3$}  }}] {};
\node[state5] (l0ts4)  [right of=l0ts3,            label={[label distance=0.05cm,text=black]below:{\huge $e_5$ \vphantom{\huge {f}$_3$}  }}] {};
\node[state6] (l0ts5)  [right of=l0ts4,             label={[label distance=0.05cm,text=black]below:{\huge $f_6$ \vphantom{\huge {f}$_3$}  }}] {};

\node[state1] (l1ts0) [above of=l0ts0] {};
\node[state2] (l1ts1) [above of=l0ts1] {};
\node[state3] (l1ts2) [above of=l0ts2] {};
\node[state4] (l1ts3) [above of=l0ts3] {};
\node[state5] (l1ts4) [above of=l0ts4] {};
\node[state6] (l1ts5) [above of=l0ts5] {};

\path[color=gray!30] [-] (l0ts0) edge node {}  (l1ts0);
\path[color=gray!30] [-] (l0ts0) edge node {}  (l1ts1);
\path[color=gray!30] [-] (l0ts1) edge node {}  (l1ts1);
\path[color=gray!30] [-] (l0ts0) edge node {}  (l1ts2);
\path[color=gray!30] [-] (l0ts1) edge node {}  (l1ts2);
\path[color=gray!30] [-] (l0ts2) edge node {}  (l1ts2);
\path[color=gray!30] [-] (l0ts0) edge node {}  (l1ts3);
\path[color=gray!30] [-] (l0ts1) edge node {}  (l1ts3);
\path[color=gray!30] [-] (l0ts2) edge node {}  (l1ts3);
\path[color=gray!30] [-] (l0ts3) edge node {}  (l1ts3);
\path[color=gray!30] [-] (l0ts0) edge node {}  (l1ts4);
\path[color=gray!30] [-] (l0ts1) edge node {}  (l1ts4);
\path[color=gray!30] [-] (l0ts2) edge node {}  (l1ts4);
\path[color=gray!30] [-] (l0ts3) edge node {}  (l1ts4);
\path[color=gray!30] [-] (l0ts4) edge node {}  (l1ts4);
\path[color=gray!30] [-] (l0ts0) edge node {}  (l1ts5);
\path[color=gray!30] [-] (l0ts1) edge node {}  (l1ts5);
\path[color=gray!30] [-] (l0ts2) edge node {}  (l1ts5);
\path[color=gray!30] [-] (l0ts3) edge node {}  (l1ts5);
\path[color=gray!30] [-] (l0ts4) edge node {}  (l1ts5);
\path[color=gray!30] [-] (l0ts5) edge node {}  (l1ts5);

\end{tikzpicture}}}
  
  \caption{ \label{fig:pia} Language model modes for generating or evaluating 6 tokens ($a,b, \dots, f)$ when subsequence length $L=3$. The numbers denote the position embeddings (P.E.). (a) Nonoverlapping (\S\ref{sec:background}).  (b) Sliding window, stride $S=1$ . Here, after the first inference pass we ignore all outputs other than the last (\S\ref{sec:background}). (c) Caching  (\S\ref{sec:PIA-caching}) where each subsequence attends to representations of the previous one. (In the next iteration, tokens $d$, $e$ and $f$ become the cache, with P.E. 1, 2 and 3, the three new tokens get P.E. 4, 5, and 6.)} 
\end{figure*}
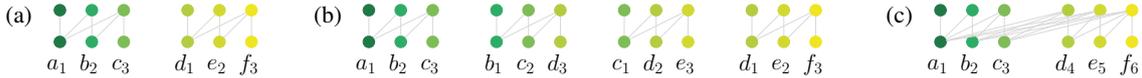

\section{Background and Experimental Setup}\label{sec:background}
Transformer language models map a list of tokens $x_{n-L:n-1}$ to a probability distribution over the next token $x_n$. We refer to the list of tokens as the \textit{current input subsequence} (whose length is $L$). Causal  masking  lets us make $L$ predictions at once, with the prediction for token $i+1$ conditioned on the $i$th token and all previous inputs $x_{n-L:i-1}$, but not on future inputs. 
We define the number of tokens the model can attend to at each timestep as its \emph{effective context window}. 
Note that $L$ is not to be confused with the (typically much greater) length of a training or evaluation dataset.

During inference, language models can be used for two distinct tasks: generation and evaluation. In order to define these tasks, we first define nonoverlapping and sliding window inference.

\paragraph{Nonoverlapping Inference}
To evaluate a string longer than $L$, we can evaluate each subsequence of $L$ tokens independently. This fast approach is commonly used during training; if used, tokens in one subsequence cannot condition on those in the previous subsequence, giving rise to the early token curse discussed in \S\ref{sec:intro}. See Figure~\ref{fig:pia}(a). 

\paragraph{Sliding Window Inference}

An alternative to the above is to use a sliding window during inference. 
Here, we choose a stride $S$ between 1 and $L-1$ and advance the window by $S$ tokens after each forward pass.\footnote{Nonoverlapping inference can be viewed as sliding window inference with stride $L$.} This means that $L-S$ tokens from the previous block are re-encoded, and only $S$ new tokens are outputted. The advantage is that all outputs in each subsequence after the first have at least $L-S$ previous tokens to condition on. However, since tokens must be re-encoded multiple times, this approach is much slower. 
When $S=1$, we output one token every inference pass, each using the maximal context window, but this is the slowest approach. 
See Figure~\ref{fig:pia}(b).

\paragraph{Minimal and Maximal Effective Context Window Sizes}
In the nonoverlapping approach, the min.  and max. effective context window sizes are $1$ and $L$, respectively. %($L_{\mathit{min}}$)
In the sliding window approach, the max. context window size is still $L$, but the min. context window size is now $L-S+1$.

\paragraph{Evaluation vs.~Generation} 
In \textit{evaluation}, a model assigns a perplexity score to a given sequence. 
Evaluation is done using either nonoverlapping inference or with a sliding window of any stride; since we already have the target sequence we can simultaneously make predictions for multiple timesteps using causal masking. 
In \textit{generation}, a model generates a new sequence, as in demonstrations of GPT-3~\cite{gpt3}. 
Generation is done only with a sliding window with stride $S=1$, which we refer to as token-by-token generation.
During generation, we append to the input a single new token, get a prediction from the model about the next token (e.g., using beam search or picking the token with the highest probability); the process is then repeated.\footnote{In this paper we do not consider open-ended generation; we generate the dev.~set, and for next-token prediction we use the ground truth token. This has the same complexity as sampling the token with the highest probability. }

\paragraph{Experimental Setup}
Our baseline is the~\citet{baevski} model, henceforth \ba, trained and evaluated on WikiText-103 \citep{pointer}. We use this baseline because of its prominent role in recent  language modeling developments~\cite{khandelwal20generalization, sandwich}. The training set contains 103.2 million tokens from  English Wikipedia.  The \ba model has $16$ transformer layers of dimension $1{,}024$, with $8$ heads in each self-attention sublayer, and feedforward sublayers with an inner dimension of $4{,}096$. This model ties the word embedding and softmax matrices~\citep{tying, inan2017} and uses sinusoidal position embeddings. It has a subsequence length of $3{,}072$ tokens and achieves a perplexity of 18.65 $\pm$ 0.24 (std. dev.) on the development set.
In our experiments, other than varying the subsequence length, we modify no other hyperparameters, including the random seed and number of training epochs (205). 
\section{How Does Context Window Size Affect Transformers?}
\label{sec:instance-size-matters}

Segmenting a corpus into subsequences results in different effective context windows for different timesteps depending on where they fall in a segment.  Subsequence length $L$ is an upper bound on the effective context window at each timestep. %%%last 3 words can be cut
When making the first prediction, the model attends only to the first input token. 
When making the second prediction, the model attends to the first two inputs, and so on, up to the $L$th timestep where the model can attend to all input tokens when making the $L$th prediction. 
\subsection{Context Window Size Matters}

\begin{table}[htbp]
\centering
\small
\centering
\setlength{\tabcolsep}{5pt}

\begin{tabular}{@{}lccccc@{}} \toprule
& Train &  \multicolumn{4}{c}{Inference} \\ \cmidrule(lr){2-2} \cmidrule(lr){3-6}
  & & \multicolumn{2}{c}{\multirow{2}[0]{*}{Nonoverlapping}} & \multicolumn{2}{c}{Sliding Window} \\ % 
\multirow{2}[4]{*}{\shortstack[l]{Subseq.\\Length}} &&&&\multicolumn{2}{c}{(Token-by-token)} \\  \cmidrule(lr){3-4} \cmidrule(lr){5-6}
 &    Speed $\uparrow$  & PPL $\downarrow$ &  Speed  $\uparrow$  & PPL $\downarrow$ & Speed $\uparrow$\\ \midrule
32   & 28.3k & 35.37 &   2.4k       &   24.98    &   \textbf{74}      \\
64   & 28.5k & 28.03 &   4.8k       &   21.47    &   69      \\
128  & \textbf{28.9k} & 23.81 &  9.2k&   19.76    &   70      \\
256  & 28.1k & 21.45 &  14.8k       &   18.86    &  63       \\
512  & 26.1k & 20.10 &  18.1k       &   18.41    &  37      \\
1024 & 22.9k & 19.11 &  \textbf{18.3k}&  17.97    &  18       \\
1536 & 18.4k & 19.05 &  17.1k       &  18.14    &  11       \\
3072 & 13.9k & \textbf{18.65} &  14.7k &   \textbf{ 17.92}    &  5       \\ \bottomrule
\end{tabular}
\caption{\label{tab:bl_contextwindow} Subsequence length's effects on performance of the \ba model on the WikiText-103 dev.~set. The baseline is the last row.  Token-by-token inf.~was computed with a sliding window stride $S=1$ to output one token at a time; see \S\ref{sec:background}. We measure speed in tok./sec.~per GPU and use a batch size of 1 for inf. }
\end{table}

Table~\ref{tab:bl_contextwindow} explores the effect of subsequence length in the \ba model on training runtime and on dev. set perplexity and runtime.\footnote{For consistency, throughout the paper we run  inference with a batch size of one. This causes models shorter  than $L=512$ to run slowly (in N.o.~eval.), although during batched N.o.~eval. they are slightly faster than the $L=512$ model.} We fix the number of tokens in each batch to $9{,}216$ but vary the subsequence length $L$ and batch size (so the product of the batch size and subsequence length remains at $9{,}216$). We report results for both nonoverlapping inference and sliding window inference with stride $S=1$, which generates only one new token per forward pass; it thus has the maximal effective context window for each generated token. 
We find that performance increases as $S$ decreases until it reaches a peak and then stops improving (not shown in Table~\ref{tab:bl_contextwindow}).\footnote{For example, the $L=3{,}072$ model's performance peaked at $S=512$ (used in~\citet{baevski}) and then stopped improving. Thus,  the result shown in Table~\ref{tab:bl_contextwindow} for that model with $S=1$ can also be achieved with $S=512$ even though that runs 500 times faster, at 2.5k tok./sec. }

We derive the following conclusions:

\textbf{Training on long sequences is expensive.} 
Models trained on subsequences of length 256 are twice as fast as models trained on subsequences of $3{,}072$ tokens, but gains for even shorter  lengths are negligible (Tab.~\ref{tab:bl_contextwindow}, col. 2).

\textbf{Long subsequence lengths can improve results.} When using the na\"{i}ve approach, nonoverlapping evaluation, we see a monotonic decrease in dev. perplexity when increasing $L$ (Tab.~\ref{tab:bl_contextwindow}, col. 3). 

\textbf{Increasing the \emph{minimum} effective context window size is more important than increasing the maximum one.} 
Using a sliding window for token-by-token evaluation 
substantially improves results for all models (Tab.~\ref{tab:bl_contextwindow}, col. 5). Here, we see negligible improvement between the models trained with subsequence lengths of $1{,}024$ and $3{,}072$ tokens (0.05 perplexity). 
This approach improves results by increasing the minimum amount of context available at each timestep which indicates that long contexts may not be beneficial to transformer models, but very short contexts are harmful. However, sliding window inference can be expensive since each token is encoded many times. For example, token-by-token inference for the $L=3{,}072$ model is almost $300$ times slower than nonoverlapping inference.

\section{Training Subsequence Length}
\label{sec:ss}
\S\ref{sec:instance-size-matters} results show that models  trained on shorter subsequences can be effective at test time, and are much faster to train. We further explore this below.

\subsection{Staged Training}

We propose a two-stage training routine that initially uses short input subsequences followed by long subsequences.\footnote{Curriculum learning \citep{CL} trains on easier inputs before progressing to harder ones. Our approach does not change the order in which the training examples are given to the model, but instead modifies their lengths.}  This method was previously applied to speed up the training of BERT~\cite{bert}, but we show 
that it also improves perplexity.

We use sinusoidal position embeddings; learned position embeddings, which we do not consider, create a dependency between the parameterization and  subsequence length. In our experiments, we neither modify nor reset the state of the optimization algorithm between the two stages.

\subsection{Experiments}
\label{sec:ss_results}

Our experimental setup is described in \S\ref{sec:background}. We do not change any hyperparameters other than reducing subsequence length while correspondingly increasing batch size to keep the number of tokens per batch constant. As in the baseline, all models are trained for 205 epochs.

All models are trained in two stages;  the second stage always uses a subsequence length of $3{,}072$, since that lead to the best performance (discussed at end of this subsection).

Appendix Table~\ref{tab:ss_match} shows the time each training routine takes to match the baseline model's performance on the validation set of WikiText-103.\footnote{Table~\ref{tab:ss_match_epoch} in the appendix shows the \emph{epoch} at which every model matched the baseline's performance.} Many configurations match this performance in less than half the time it takes to train the baseline itself; some reach baseline performance in only 37\% of the time needed to train the baseline.

Although all models take less time to train than the baseline, Table~\ref{tab:ss_best} shows that many \textbf{outperform} it. For example, the best model---trained with subsequence length $L=128$ until epoch 50---outperforms the baseline by 1.1 perplexity despite completing training in 87\% of the time the baseline takes to do so.  The model that trains with $L=128$ until epoch 100 achieves similarly strong results (17.62 perplexity) and finishes training in 74\% of the time it takes the baseline.\footnote{Table~\ref{tab:ss_total_time} in the appendix shows the total time it took to train each model.}

\begin{table}[t]

\newcommand*{\MidNumberzero}{17.525}
\newcommand*{\MidNumberone}{17.785} 
\newcommand*{\MidNumbertwo}{18.06} 
\newcommand*{\MidNumberthree}{18.335}  
\newcommand*{\MaxNumber}{18.63}%  

\newcommand{\ApplyGradient}[1]{%
        \ifdim #1 pt < \MaxNumber pt\relax%
        \ifdim #1 pt < \MidNumberthree pt\relax%
        \ifdim #1 pt < \MidNumbertwo pt\relax%
        \ifdim #1 pt < \MidNumberone pt\relax% 
        \ifdim #1 pt < \MidNumberzero pt\relax
        
           \colorbox{mycolor0!70}{\textbf{#1}}
        \else

           \colorbox{mycolor0!70}{#1}
        \fi
        \else% %%%%for values between midnumber 2 and midnumber 1 color in color 3 (l3rd ightest color)

            \colorbox{mycolor1!70}{#1}
        \fi%
        \else%  %%for values between midnumber 3 and midnumber 2 color in color 2 (2nd lightest color)

            \colorbox{mycolor2!70}{#1}
        \fi%
        \else% %%%for values between max and midnumber 3 color in color 3 (lightest color)

            \colorbox{mycolor3!70}{#1}
        \fi%
        \else %%for larger values than max color white

            \colorbox{white}{#1}
        \fi
}

\newcolumntype{R}{>{\collectcell\ApplyGradient}c<{\endcollectcell}}

\small
\centering
\setlength{\tabcolsep}{-0pt}

\begin{tabular}{@{}l@{\hspace{4pt}}l@{\hspace{4pt}}|RRRRRRR@{}}

\toprule
& &\multicolumn{7}{c@{}}{\text{Initial Stage Subseqence Length}}\\

&     & 32    & 64    & 128   & 256   & 512   & 1024  & 1536  \\ \specialrule{\lightrulewidth}{0pt}{0pt}
& 25  & 17.94 & 17.57 & 17.58 & 18.19 & 18.06 & 18.20 & 18.77 \\
& 50  & 17.81 & 17.59 & 17.52 & 18.08 & 18.01 & 18.14 & 18.62 \\
& 75  & 17.93 & 17.61 & 17.55 & 18.01 & 18.05 & 18.03 & 18.57 \\
\multirow{2}{*}{\smash{\rotatebox[origin=c]{90}{\text{Initial Stage Epochs}}}}
& 100 & 18.14 & 17.67 & 17.62 & 18.00 & 18.10 & 18.00 & 18.51 \\
& 125 & 18.61 & 17.88 & 17.70 & 18.00 & 18.13 & 17.98 & 18.49 \\
& 150 & 19.45 & 18.37 & 17.98 & 18.01 & 18.15 & 18.00 & 18.49 \\
& 175 & 21.16 & 19.51 & 18.57 & 18.23 & 18.20 & 18.08 & 18.57 \\
& 200 & 35.38 & 28.03 & 23.80 & 21.45 & 19.63 & 18.56 & 18.84 \\ \bottomrule
\end{tabular}
\caption{\label{tab:ss_best} Each model's perplexity at the end of training (dev.~set, nonoverlapping eval.). All models have a subsequence length of $3{,}072$ tokens at the end of training. The \ba baseline  achieves 18.65 $\pm$ 0.24 perplexity. }
\end{table}

These results are very robust to the choice of initial stage subsequence length and number of epochs. Table~\ref{tab:ss_best} shows that all models with an initial stage of $L = 1{,}024$ tokens or less that switch to the second stage at epoch 125 or before beat the baseline by a large margin at the end of training. Additionally, Appendix Table~\ref{tab:ss_match} shows that those models match the baseline's perplexity in at most 71\% of the time it takes to train the baseline. 

\begin{figure*}[t!]
\centering

  \sidecaption{subfig:a}
  \raisebox{-\height}{\includegraphics[scale=0.73,valign=t]{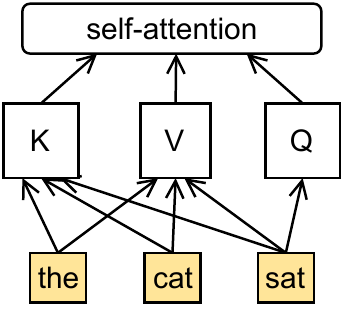} 
    \vphantom{ \includegraphics[scale=0.73,valign=t]{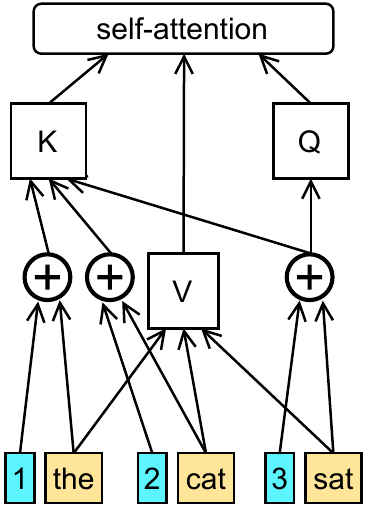} }}
  \qquad \qquad \qquad
    \sidecaption{subfig:b}
  \raisebox{-\height}{\includegraphics[scale=0.73,valign=t]{images/pia-after2.pdf} 
   }

\caption{\label{fig:pia-noah-version}
Inputs to the self-attention sublayer, conventionally (left)  and with position-infused attention (right), for $L=3$, at timestep $3$. The numbers denote the position embeddings. } 
\end{figure*}

When we use nonoverlapping evaluation, the \ba baseline obtains 18.65 perplexity on the development set; our best model obtains 17.52.
When we use sliding window evaluation (following Baevski \& Auli, we use stride $S=512$), our best model obtains 16.89 perplexity, a large improvement on the 17.92 \ba result  in that setting. On the test set, using the same sliding window evaluation, our model obtains 17.56 perplexity, a substantial gain over the baseline's 18.70 test-set perplexity. Appendix Table~\ref{tab:combined_mem} shows that our best model uses almost five times less memory during the first stage than the baseline.

We also found that setting $L$ to less than $3{,}072$ tokens in the second stage degraded performance. (Appendix Table~\ref{tab:ss_l128e50} shows staged training results with an initial stage length of 128 for 50 epochs (as in the best model) and varying lengths for the second stage. We found this to also be true for other initial stage lengths and epochs.) Unlike results in Table~\ref{tab:bl_contextwindow}, where we show that models with $L$ larger than $1{,}024$ do not substantially improve token-by-token generation perplexity, models trained using staged training improve when given longer inputs (Appendix Table~\ref{tab:ss_l128e50}).
Further, we explored using more than two stages (up to six), but this did not outperform our two-stage curriculum.  

Finally, Appendix~\ref{sec:books}  shows that staged training  substantially improves on the Toronto Book Corpus~\cite{zhu2015aligning}.

\section{Repositioning Position Embeddings}
\label{sec:pia}

Sliding window inference substantially improves performance by increasing the minimum effective context window size. 
But it is very slow. 
We could solve this by letting the model attend to representations of the previous subsequence during inference on the current one.

In this case, the same token representations would be used in different positions since a token generated near the end of one subsequence would be cached and reused near the start of the next one. However, transformer model representations entangle positional and content information, so a cached token representation would encode an incorrect position when reused in a new position. 

TransformerXL \citep{transformer-xl} uses \emph{relative} position embeddings to solve this problem. However, that approach is slower and uses more parameters and memory than the baseline transformer.\footnote{The self-attention coefficients between $q$ queries and $k$ keys  in TransformerXL are the sum of two dot products of size $q\cdot k$; the unmodified attention sublayer and our PIA method both compute only  one dot product of size $q\cdot k$. We also benchmarked the TransformerXL model using its publicly released code and found that their relative position embeddings slow inference by 22\% and require 26\% more parameters than their implementation of the unmodified self-attention sublayer. } 

We solve this using no extra parameters, memory, or runtime. We also show that our method can use much \emph{shorter} input subsequences and still achieve superior performance.

\paragraph{Transformer Language Models}

The baseline transformer LM, given a token list $T$ of length $L$ and a tensor $\mathbf{P}$ containing the first $L$ position embeddings, produces $L$ next-token predictions using the following procedure:

\begin{enumerate}
    \item Embed each token in ${T}$, producing tensor $\mathbf{X}$.  
    \item \label{step:pos} Add the position embedding of each index to the token at that index: $\mathbf{X} = \mathbf{X} + \mathbf{P}$.
    \item Feed $\mathbf{X}$ through each transformer layer. The self-attention sublayer in each transformer layer is invoked as follows: 
    $\textrm{self-attention}(\textrm{key}{=}\mathbf{X}, \textrm{query}{=}\mathbf{X}, \textrm{value}{=}\mathbf{X})$
    \item Transform the outputs of the last transformer layer using the softmax layer, giving $L$ next-token probability distributions. 
\end{enumerate}

\subsection{Position-Infused Attention (PIA)}

We propose to let the model reuse previous outputs by making each output contain no explicit positional information. 
To do this, we modify the model so that it does not add position embeddings at the \emph{beginning} of the computation (step~\ref{step:pos}), but rather adds them to the query and key vectors at each layer (but \emph{not} to the value vectors). The outputs at each layer are the transformed, weighted sums of the value vectors, and, since the value vectors in our model do not contain explicit positional information, the outputs also do not.

Formally, steps 1 and 4 do not change, step 2 is omitted, and step 3 is modified to invoke the self-attention sublayer as follows:
\begin{align*}
\textrm{self-attention}(
&\textrm{key}{=}\mathbf{X}{+}\mathbf{P}, \textrm{query}{=}\mathbf{X}{+}\mathbf{P}, \\  
&\textrm{value}{=}\mathbf{X}) 
\end{align*}
Figure~\ref{fig:pia-noah-version} (b) depicts  this method.

Although PIA sublayer outputs contain no explicit positioning information, the attention mechanism can still compute position-dependent outputs because positional information is added to the query and key vectors. Our method is implementable in just a few lines of code.

\subsection{PIA Enables Caching} \label{sec:PIA-caching}

In the unmodified transformer, to generate a string whose length exceeds $L$, it would have to be split into separate subsequences, and the model would  be unable to attend to the previous subsequence when generating the current one. 

Using PIA, we can store and attend to representations of the previous subsequence since they no longer contain any explicit positioning information. 

Therefore, all our PIA models use a cache, where representations from the previous forward pass are stored and attended to in the next forward pass.

\paragraph{Caching makes generation faster.}
The complexity of the attention mechanism is $O(q \cdot k)$ where $q$ is the number of queries (outputs) and $k$ is the number of key-value pairs (inputs). To  generate a sequence whose length exceeds $L$ using token-by-token generation in the unmodified transformer (with subsequence length $L$), attention takes $O(L^2)$ time (since there are $L$ queries and $L$ keys). Using PIA and caching, we can reuse $L-1$ of the previous outputs at every layer. Thus, our attention sublayer takes $O(L)$ time (because now there is a single query and $L$ keys).

Our approach is useful in scenarios where we need to evaluate or generate sequences that are longer than the model's subsequence length. Therefore, it would not be applicable to sequence-to-sequence tasks such as sentence-level translation, where sequence lengths are short.

Most language models, including \ba, train on their data as nonoverlapping subsequences.
This means that training subsequences can be shuffled at each epoch and consumed in random order. However, when using PIA, we would like the cache to contain the previous subsequence. We therefore do not shuffle the data, making the cached subsequence the previously occurring one.

Figure~\ref{fig:pia}(c) depicts training with a cache that contains 
representations of the previous subsequence.

\subsection{Experiments} \label{sec:pia_results}

We use the experimental setup described in \S\ref{sec:background}.

The \ba baseline achieves 18.65 on the development set.
We train two additional baselines, the first uses PIA without caching and the second uses caching but no PIA.
If just PIA is used (without caching), performance degrades to 19.35 perplexity, but the model's speed and memory usage do not change. Using caching without PIA severely hurts performance, obtaining 41.59 perplexity. 
Disabling data shuffling in the PIA-only model achieves similar performance to that model when it does use data shuffling, at 19.44 perplexity. Not shuffling the data is necessary for recurrent-style training that caches previously computed subsequence representations.

Our next experiments use the recurrent-style training of~\citet{transformer-xl}, where we receive $L$ new tokens at every training iteration and attend to $L'$ cached representations (of the subsequence of tokens that came immediately prior to the $L$ new tokens). As before, we output $L$ predictions at every training iteration.
This means that the maximal and minimal effective context window sizes are $L'+L$ and $L'+1$, respectively.

In all our models with PIA and caching, we set $L'=L$ because a manual exploration of different models where $L' \neq L$ did not yield better results.

Table~\ref{tab:pia_bestperp} compares the results of our models that use PIA and caching to the baseline on the WikiText-103 dev.~set. Evaluation and generation speeds are shown in the nonoverlapping (N.o.) and sliding window (S.W., with stride $S=1$) speed columns, respectively.\footnote{Note that~\citet{baevski} show that the baseline model can also achieve 17.92 during S.W. evaluation, when $S=512$, with a speed of 2.5k tokens per second.}
Unlike in the baseline, token-by-token evaluation in our model achieves the same perplexity as nonoverlapping evaluation since in both cases, the predictions for each input subsequence are conditioned not only on the current input, but also on the previous input, making the context window the same in both inference modes (in both cases, at every timestep, the context window is all tokens up to that timestep).

\begin{table}[]
\small
\centering

\begin{tabular}{@{}lcccc@{}}
\toprule

& Train & \multicolumn{3}{c}{Inference} \\  \cmidrule(lr){2-2} \cmidrule(lr){3-5}
\multirow{2}[3]{*}{\shortstack[l]{Subseq.\\ Length}}& & &  \multicolumn{2}{c}{Speed  $\uparrow$} \\ \cmidrule(lr){4-5}

& Speed $\uparrow$    &  PPL  $\downarrow$  &  N.o. & S.W. \\ \midrule
    32   &   22.0k            &20.53                    &2.0k & 49   \\ 
    64   &   23.8k            &19.07                    &4.1k & \textbf{51}   \\
    128  &   \textbf{24.4k}   &18.37                    &7.9k & 50   \\          
    256  &   23.5k            &17.92                    &12.8k& 48   \\
    512  &   21.5k            &\textbf{17.85}           &14.5k& 46   \\
    768  &   17.6k            &18.16                    &13.8k& 43   \\
    1024 &   16.6k            &18.19                    &13.9k& 39   \\
    1536 &   12.9k            &19.11                    &7.9k & 34   \\ \midrule
    Baseline& \multirow{2}{*}{13.9k}   &  18.65  &\textbf{14.7k} & -  \\
    (3072) &                     &  17.92 & -  &5 \\
    \bottomrule
    
\end{tabular}
\caption{\label{tab:pia_bestperp} Dev.~perplexity and speed for PIA models trained with different subsequence lengths ($L$). PIA models attend to $L$ new and $L$ cached tokens at each inference pass. 
N.o. is nonoverlapping eval.; S.W. is sliding window eval., where we always use $S=1$ (token-by-token) here. The baseline is evaluated with both evaluation methods. 
We measure speed in tok./sec.~per GPU and use a batch size of 1 for inference. }
\end{table}

Table~\ref{tab:pia_bestperp} shows that as we increase subsequence length, perplexity improves, peaking at $512$ before starting to degrade. 
Our best model obtains 17.85 perplexity, which is multiple standard deviations better than the baseline (18.65, N.o.). Table~\ref{tab:combined_test} in \S\ref{sec:combined} shows a similar gain on the test set. The best model runs 1\% slower than the baseline during N.o. eval.~(since caching reduces the speed gain from smaller attention matrices in this mode). Table~\ref{tab:combined_mem} (appendix) shows that it uses less than half of the memory the baseline does during training. Our best model trains 55\% faster than the baseline. 

Our best model, with subsequence length $512$, has attention matrices of size $512 \cdot 1{,}024$ (since we have $512$ queries---one per every new token---and $1{,}024$ keys and $1{,}024$ values---one per every new token and every cached token). In the baseline, all attention matrices are of size $3{,}072 \cdot 3{,}072$.

Caching previously computed representations lets us 
do token-by-token generation efficiently when generating more than $L$ tokens.
Our model is nine times faster than the baseline at token-by-token generation even as it achieves better perplexity and uses much less memory (Tab.~\ref{tab:pia_bestperp}, col. 5).

PIA and caching also greatly improve perplexity on the Toronto Book Corpus; see~\ref{sec:books} in the appendix.

\section{Shortformer Results}
\label{sec:combined}

\begin{table}[]
\small
\centering

\begin{tabular}{@{}lccc@{}}
\toprule
\multirow{2}[3]{*}{\shortstack[l]{First Stage\\Subseq. Length}} & Train & Inference \\ \cmidrule(lr){2-2} \cmidrule(lr){3-3}
 & Speed $\uparrow$ & PPL  $\downarrow$  \\ \midrule
    32      &   21.6k   & 17.66 \\
    64      &   22.6k   &  17.56 \\
    128     &   \textbf{22.9k}   & \textbf{17.47} \\
    256     &   22.5k   & 17.50 \\ \midrule
PIA + Cache w/o     &  \multirow{2}{*}{21.5k}   &  \multirow{2}{*}{17.85} & \\ 
Staged Training &  & \\\bottomrule
    
\end{tabular}
\caption{\label{tab:combined} Dev.~perplexity for models that use PIA, caching, and staged training (with final subseq. length of 512). We measure speed in tok./sec.~per GPU. Evaluation speed is the same for all models, at 14.5k tok./sec.}
\end{table}

To assess whether the gains from staged training, PIA and caching are additive, we take our best caching PIA model, with subsequence length $512$, and apply staged training to it,  training it with a subsequence length of between 32 to 256 for the first half of training.\footnote{We picked $50\%$ of epochs as the length of the first stage since that produced near-optimal results at a fast speed in \S\ref{sec:ss}.} Table~\ref{tab:combined} shows the results. As in \S\ref{sec:ss_results}, where staged training was applied to the unmodified baseline, the results are very robust to the choice of initial stage subsequence length, with \emph{all} the different choices improving perplexity over the model that does not use staged training. 

The best model (with initial subsequence length 128), which we call Shortformer, achieves 17.47 dev.~set perplexity and trains 65\% faster than the baseline. Since its attention matrices are of dimension $512 \cdot 1{,}024$ (the baseline's are $3{,}072 \cdot 3{,}072$), our model uses less memory (\S\ref{sec:mem}, appendix). It has the same number of parameters as the baseline.

Figure~\ref{fig:all} (appendix) compares our best models using each method we presented (and their combination) to the baseline. It shows that combining caching, PIA and staged training (Shortformer) yields the quickest training and  best perplexity when using nonoverlapping evaluation. Evaluation speed is similar for all of these models.

Finally, Table~\ref{tab:combined_test} compares our best models on the test set of WikiText-103 to the state of the art.\footnote{We benchmarked speed, on V100 GPUs, for all models that had publicly available code. }

\begin{table}[t]
\centering
\small
\setlength{\tabcolsep}{2pt}
\begin{tabular}{@{}lccccc@{}} \toprule
&&Train& \multicolumn{3}{c}{Inference (Test)} \\  \cmidrule(lr){3-3}\cmidrule(lr){4-6}
Model  & Param. $\downarrow$& Speed $\uparrow$& Mode & Speed $\uparrow$  & PPL  $\downarrow$ \\ \midrule
\multirow{2}{*}{Baseline}&\multirow{2}{*}{\textbf{247M}}  & \multirow{2}{*}{13.9k} & N.o. & \textbf{14.7k}& 19.40 \\
&&&S.W.&2.5k&18.70\\
TransformerXL$^*$ &257M & 6.0k & N.o.& 3.2k & 18.30 \\
Sandwich T.       &\textbf{247M}& 13.9k&S.W. &2.5k & 17.96 \\
Compressive T.    &329M         &-&N.o.&-&17.1\\
Routing T.        & - & - & N.o& - &15.8 \\
kNN-LM$^{**}$     &\textbf{247M}& 13.9k&S.W. & 145 & \textbf{15.79} \\ \midrule
PIA + Caching  &\textbf{247M}  & 21.5k & N.o. & 14.5k& 18.55   \\
Staged Training   &\textbf{247M}&17.6k&S.W.&2.5k &17.56 \\
Shortformer       &\textbf{247M}& \textbf{22.9k}& N.o.& 14.5k& 18.15    \\ \bottomrule

\end{tabular}
\caption{\label{tab:combined_test} Comparison of our best models to other strong LMs (see text for citations and explanations) evaluating the WikiText-103 test set, where $S=512$. We measure speed in tok./sec.~per GPU, and use a batch size of 1 for inference.  $^*$TransformerXL runs on an older version of PyTorch, which might affect  speed. $^{**}$kNN-LM requires a 400GB datastore.}
\end{table}

Shortformer is almost twice as fast to train as the baseline and achieves superior results. Like the best model from \S\ref{sec:pia_results}, it is nine times faster than the baseline for token-by-token generation.

Since it uses a cache, sliding window evaluation does not increase Shortformer's performance. By training the baseline with staged training (and no PIA or caching), we obtain a model (our best model from \S\ref{sec:ss_results}) that, with sliding window eval., obtains even better results, but that model is much slower than Shortformer (Table~\ref{tab:combined_test}, second-to-last row).

Shortformer outperforms the baseline's perplexity and performs within a standard deviation of the Sandwich Transformer~\cite{sandwich} and TransformerXL. It does not outperform the Compressive Transformer~\citep{Rae2020Compressive}, Routing Transformer~\cite{roy2020efficient} and kNN-LM~\citep{khandelwal20generalization}, which make orthogonal improvements that can be applied to any language model, at the price of slower decoding. Combining them with our approach may yield further gains. These results are similar to those we obtain on the Toronto Book Corpus (\S\ref{sec:books} in the appendix). 

\section{Related Work}

\paragraph{Staged Training}

\citet{bert} used a staged training routine for BERT by performing the first 90\% of training on short subsequences (of length $128$) before moving on to longer ones (of length $512$). 
They use this method to speed training, but we show that also it improves perplexity and analyze different configurations of this method. 

Many recent papers have explored improving transformer efficiency by reducing the quadratic cost of self-attention, motivated by scaling to longer sequences \cite{kitaev2020reformer, roy2020efficient, tay2020efficient}. We  
instead demonstrate improved results with shorter sequences, which naturally also improve efficiency.

One way to reduce transformer memory usage is to sparsify the attention matrix by letting the model attend only to a subset of nearby tokens at each timestep \cite{sparse-transformer, longformer, roy2020efficient}. Training on shorter subsequence lengths is much more efficient: we use multiple, but much smaller, attention matrices.  
Since attention uses memory and computation in a way that scales quadratically with input size, splitting the inputs into multiple subsequences each processed independently lets us use less memory and run faster. 
Like our method,~\citet{longformer} attend at each timestep to a growing number of neighbors as training progresses, but they use five stages, which we found not to be superior to our two-staged method. 

The adaptive attention span model of~\citet{sukhbaatar2019adaptive} learns the maximum effective context window sizes for each head at each layer independently. Like in our method,  context window sizes are smaller at the start of training and lengthen as training progresses. We show that a simple approach of manually choosing two subsequence lengths is highly effective. In addition, keeping subsequence lengths equal across all heads and layers lets us save memory and runtime. 
\paragraph{Position-Infused Attention}

TransformerXL \cite{transformer-xl} caches and attends to previous representations using an attention sublayer that uses relative positioning~\cite{shaw}. It runs much slower than the unmodified attention sublayer, requires extra parameters, and requires internally modifying the self-attention sublayer, while our PIA method (\S\ref{sec:pia}) does not. 

In parallel with our work,~\citet{ke2020rethinking} compute  attention coefficients by summing two attention matrices, one based on position-position interactions and the other on content-content interactions. As in PIA, they do not add position embeddings at the bottom of the model.
They present results only for BERT, which uses much smaller subsequences than our models. 

\section{Conclusion}

Our results challenge the conventional wisdom that longer subsequences are \emph{always} better. % for transformers. %%can remove last 2 words
By first training on shorter subsequences and then progressing to longer ones via staged training, we improve perplexity and reduce training time. 
We additionally propose position-infused attention, which enables caching and efficiently attending to previous outputs; we show that models using this method do not require large input subsequences.
We finally show that these two methods can be combined to produce a speedier and more accurate  model. 

\section*{Acknowledgments}
We thank Tim Dettmers, Jungo Kasai, Gabriel Ilharco, Hao Peng, Sewon Min,  Mandar Joshi, Omer Levy, Luke Zettlemoyer, Julian Michael, Edward Misback, Sofia Serrano, Nikolaos Pappas, Jesse Dodge, Myle Ott, and Sam Shleifer for their valuable feedback and fruitful discussions.

\bibliographystyle{acl_natbib}
\bibliography{acl2021}

\pagebreak 
\clearpage
\appendix
\section{Appendix}

\subsection{Additional Staged Training Results}
Table~\ref{tab:ss_match} shows the time each staged training model needs to match baseline performance, as a fraction of the time it takes to train the baseline. The fastest three configurations each match the baseline's performance in just 37\% of the time it takes to train the baseline. This result is very robust to hyperparameter changes, as all models trained with initial subsequence length of between 64 and 512, that switch to the second stage at epoch 50 to 150, manage to match the baseline's performance in at most 59\% of the time it takes to train it.

\begin{table}[htp]

\newcommand*{\minmin}{0.1}
\newcommand*{\MidNumberzero}{0.375}
\newcommand*{\MidNumberone}{0.4} %  
\newcommand*{\MidNumbertwo}{0.5} %  
\newcommand*{\MidNumberthree}{0.6} %  
\newcommand*{\MaxNumber}{0.7}% %white

\newcommand{\ApplyGradient}[1]{%
        \ifdim #1 pt < \MaxNumber pt\relax%
        \ifdim #1 pt < \MidNumberthree pt\relax%
        \ifdim #1 pt < \MidNumbertwo pt\relax%
        \ifdim #1 pt < \MidNumberone pt\relax% 
        \ifdim #1 pt < \MidNumberzero pt\relax
        \ifdim #1 pt < \minmin pt\relax
        
            {}
        \else
        
           \colorbox{mycolor0!70}{\textbf{#1}} 
        \fi
        \else

           \colorbox{mycolor0!70}{#1}
        \fi
        \else% %%%%for values between midnumber 2 and midnumber 1 color in color 3 (l3rd ightest color)

            \colorbox{mycolor1!70}{#1}
        \fi%
        \else%  %%for values between midnumber 3 and midnumber 2 color in color 2 (2nd lightest color)

            \colorbox{mycolor2!70}{#1}
        \fi%
        \else% %%%for values between max and midnumber 3 color in color 3 (lightest color)

            \colorbox{mycolor3!70}{#1}
        \fi%
        \else %%for larger values than max color white

            \colorbox{white}{#1}
        \fi
}

\newcolumntype{R}{>{\collectcell\ApplyGradient}c<{\endcollectcell}}

\small
\centering
\setlength{\tabcolsep}{-0pt}

\begin{tabular}{@{}l@{\hspace{4pt}}l@{\hspace{4pt}}|RRRRR@{\hspace{-1pt}}R@{\hspace{-2pt}}R@{}}

\toprule
& &\multicolumn{7}{c@{}}{\text{Initial Stage Subsequence Length}}\\

& 0   & 32    & 64    & 128   & 256   & 512   & 1024  & 1536  \\ \specialrule{\lightrulewidth}{0pt}{0pt}
& 25  & 0.60 & 0.54 & 0.53 & 0.65 & 0.64 & 0.71 &   0  \\
& 50  & 0.53 & 0.48 & 0.47 & 0.54 & 0.59 & 0.63 & 0.81 \\
& 75  & 0.51 & 0.43 & 0.42 & 0.48 & 0.53 & 0.56 & 0.79 \\
\multirow{2}{*}{\smash{\rotatebox[origin=c]{90}{\text{Initial Stage Epochs}}}} 
& 100 & 0.52 & 0.40 & 0.38 & 0.41 & 0.47 & 0.50 & 0.73 \\
& 125 & 0.61 & 0.41 & 0.37 & 0.37 & 0.42 & 0.46 & 0.69 \\
& 150 &  0    & 0.48 & 0.39 & 0.37 & 0.40 & 0.44 & 0.66 \\
& 175 &   0   &   0   & 0.48 & 0.43 & 0.45 & 0.51 & 0.70 \\
& 200 &    0  &    0  & 0     &  0    &  0    & 0.59 &  0    \\ \bottomrule
\end{tabular}
\caption{\label{tab:ss_match} Time needed to match baseline performance (dev.~set, nonoverlapping eval.) as a fraction of time needed to train the baseline (smaller is better). Models never matching the baseline have empty cells. All models have a subsequence length of $3{,}072$ tokens at the end of training.} 
\end{table}

\begin{table}[htp]
\newcommand*{\minmin}{0.1}
\newcommand*{\MidNumberzero}{122.5}
\newcommand*{\MidNumberone}{132} % 
\newcommand*{\MidNumbertwo}{142} %  
\newcommand*{\MidNumberthree}{152} %  
\newcommand*{\MaxNumber}{162}% %white

\newcommand{\ApplyGradient}[1]{%
        \ifdim #1 pt < \MaxNumber pt\relax%
        \ifdim #1 pt < \MidNumberthree pt\relax%
        \ifdim #1 pt < \MidNumbertwo pt\relax%
        \ifdim #1 pt < \MidNumberone pt\relax% 
        \ifdim #1 pt < \MidNumberzero pt\relax
        \ifdim #1 pt < \minmin pt\relax
        
            {}
        \else
        
           \colorbox{mycolor0!70}{\textbf{#1}}
        \fi
        \else

           \colorbox{mycolor0!70}{#1}
        \fi
        \else% %%%%for values between midnumber 2 and midnumber 1 color in color 3 (l3rd ightest color)

            \colorbox{mycolor1!70}{#1}
        \fi%
        \else%  %%for values between midnumber 3 and midnumber 2 color in color 2 (2nd lightest color)

            \colorbox{mycolor2!70}{#1}
        \fi%
        \else% %%%for values between max and midnumber 3 color in color 3 (lightest color)

            \colorbox{mycolor3!70}{#1}
        \fi%
        \else %%for larger values than max color white

            \colorbox{white}{#1}
        \fi
}

\newcolumntype{R}{>{\collectcell\ApplyGradient}c<{\endcollectcell}}

\small
\centering
\setlength{\tabcolsep}{-0pt}

\begin{tabular}{@{}l@{\hspace{4pt}}l@{\hspace{4pt}}|RRRRR@{\hspace{-2pt}}R@{\hspace{-2pt}}R@{}}

\toprule
& &\multicolumn{7}{c@{}}{\text{Initial Stage Subsequence Length}}\\

&     & \multicolumn{1}{c}{32}    & \multicolumn{1}{c}{64}    & \multicolumn{1}{c}{128}   & 256   & 512   & 1024  & 1536  \\ \specialrule{\lightrulewidth}{0pt}{0pt}
& 25  & 136 & 123 & 122 & 146 & 144 & 155 &  0   \\
& 50  & 135 & 124 & 122 & 136 & 144 & 149 & 179 \\
& 75  & 143 & 128 & 125 & 136 & 144 & 145 & 181 \\
\multirow{2}{*}{\smash{\rotatebox[origin=c]{90}{\text{Initial Stage Epochs}}}} 
& 100 & 158 & 135 & 130 & 136 & 145 & 142 & 175 \\
& 125 & 190 & 149 & 141 & 140 & 146 & 144 & 174 \\
& 150 &  0   & 176 & 160 & 154 & 153 & 151 & 174 \\
& 175 &  0   &  0   & 191 & 178 & 177 & 176 & 189 \\
& 200 &  0   &  0   &  0   & 0 &0 & 202 &  0    \\\bottomrule
\end{tabular}
\caption{\label{tab:ss_match_epoch} Epoch at which each model matches the baseline. Some models never match the baseline, and so those cells are empty. }
\end{table}

\begin{table}[htbp!]
\newcommand*{\minmin}{0.1}
\newcommand*{\MidNumberzero}{0.485}
\newcommand*{\MidNumberone}{0.6} %  
\newcommand*{\MidNumbertwo}{0.7} %  
\newcommand*{\MidNumberthree}{0.8} % 
\newcommand*{\MaxNumber}{0.9}% %white

\newcommand{\ApplyGradient}[1]{%
        \ifdim #1 pt < \MaxNumber pt\relax%
        \ifdim #1 pt < \MidNumberthree pt\relax%
        \ifdim #1 pt < \MidNumbertwo pt\relax%
        \ifdim #1 pt < \MidNumberone pt\relax% 
        \ifdim #1 pt < \MidNumberzero pt\relax
        \ifdim #1 pt < \minmin pt\relax
        
            {#1}
        \else
        
           \colorbox{mycolor0!70}{\textbf{#1}}
        \fi
        \else

           \colorbox{mycolor0!70}{#1}
        \fi
        \else% %%%%for values between midnumber 2 and midnumber 1 color in color 3 (l3rd ightest color)

            \colorbox{mycolor1!70}{#1}
        \fi%
        \else%  %%for values between midnumber 3 and midnumber 2 color in color 2 (2nd lightest color)

            \colorbox{mycolor2!70}{#1}
        \fi%
        \else% %%%for values between max and midnumber 3 color in color 3 (lightest color)

            \colorbox{mycolor3!70}{#1}
        \fi%
        \else %%for larger values than max color white

            \colorbox{white}{#1}
        \fi
}

\newcolumntype{R}{>{\collectcell\ApplyGradient}c<{\endcollectcell}}

\small
\centering
\setlength{\tabcolsep}{-0pt}

\begin{tabular}{@{}l@{\hspace{4pt}}l@{\hspace{4pt}}|RRRRR@{\hspace{-1pt}}R@{\hspace{-2pt}}R@{}}

\toprule
& &\multicolumn{7}{c@{}}{\text{Initial Stage Subsequence Length}}\\

&     & 32    & 64    & 128   & 256   & 512   & 1024  & 1536  \\ \specialrule{\lightrulewidth}{0pt}{0pt}

&25  & 0.94 & 0.94 & 0.94 & 0.94 & 0.94 & 0.95 & 0.97 \\
&50  & 0.87 & 0.87 & 0.87 & 0.87 & 0.88 & 0.90 & 0.94 \\
&75  & 0.81 & 0.81 & 0.81 & 0.81 & 0.82 & 0.85 & 0.90 \\
\multirow{2}{*}{\smash{\rotatebox[origin=c]{90}{\text{Initial Stage Epochs}}}} 
&100 & 0.75 & 0.75 & 0.74 & 0.75 & 0.77 & 0.80 & 0.87 \\
&125 & 0.68 & 0.68 & 0.68 & 0.69 & 0.71 & 0.75 & 0.84 \\
&150 & 0.62 & 0.62 & 0.61 & 0.62 & 0.65 & 0.70 & 0.81 \\
&175 & 0.56 & 0.56 & 0.55 & 0.56 & 0.59 & 0.66 & 0.78 \\
&200 & 0.49 & 0.49 & 0.48 & 0.50 & 0.53 & 0.61 & 0.75 \\ \bottomrule
\end{tabular}
\caption{\label{tab:ss_total_time} Total time needed to train each model as a fraction of the time needed for baseline training. }
\end{table}

Tables~\ref{tab:ss_match_epoch} and~\ref{tab:ss_total_time} show the epoch at which each model matched the baseline's performance and the total time it took to train each of our staged training models.

\subsection{Staged Training with Shorter Final Stage $L$}
In section~\ref{sec:ss}, all models presented used Staged Training with a final input subsequence length $L$ of $3{,}072$ tokens. In Table~\ref{tab:ss_l128e50}, we show the results of training with a first stage with $L=128$ for 50 epochs, and using varying subsequence lengths for the second stage. The best result is obtained when the second stage uses $L=3{,}072$. In addition, in all of our other experiments (not presented here) with different $L$ and epoch number values for the first stage, we observed that using $L=3{,}072$ for the second stage always achieved the best perplexities. 
Models trained with staged training and evaluated with a sliding window sometimes perform slightly worse when $S$ is decreased, but this difference is much smaller than the standard deviation. The $L=1536$ and $L=3072$ models peaked at $S=512$, and then as $S$ was decreased perplexity started slightly degrading.\footnote{We conjecture that this is because of a train-test mismatch that occurs since the average effective context length during training is $\frac{3{,}072}{2}=1{,}536$ and so the model focuses on learning how to make predictions for the tokens in the center of the input, and does not perform as well when making predictions for tokens at the end of the input (which is what we use when using sliding window evaluation). }

\begin{table}[htpb!]
\small
\centering

\begin{tabular}{@{}lccc@{}} \toprule

\multirow{3}[3]{*}{\shortstack[l]{Final\\Subseq.\\Length}}   &   \multicolumn{3}{c}{Inference PPL $\downarrow$}     \\ \cmidrule(lr){2-4}
     &\multirow{2}[0]{*}{Nonoverlapping}  &\multicolumn{2}{c}{Sliding Window}  \\ 
    &   & $S=512$ & $S=1$ \\ \midrule

256   & 21.26         &- &18.72   \\
512   & 19.69         &19.69 &18.04   \\
1024  & 18.64         &17.60 &17.58   \\
1536  & 18.10         & 17.28 &17.30   \\
3072  & \textbf{17.52}&\textbf{16.89} & \textbf{17.01}   \\ \bottomrule
\end{tabular}
\caption{\label{tab:ss_l128e50} Inference perplexity for staged training models trained with an initial stage subsequence length of $128$ for 50 epochs and varying second stage subsequence length $L$ (for the second stage's 155 epochs). $S$ is stride. To see how these models perform without staged training, refer to Table~\ref{tab:bl_contextwindow}.  }  
\end{table}

\subsection{Training Speed vs. Performance}
\begin{figure}[h!] 
\centering
\includegraphics[width=0.9\linewidth]{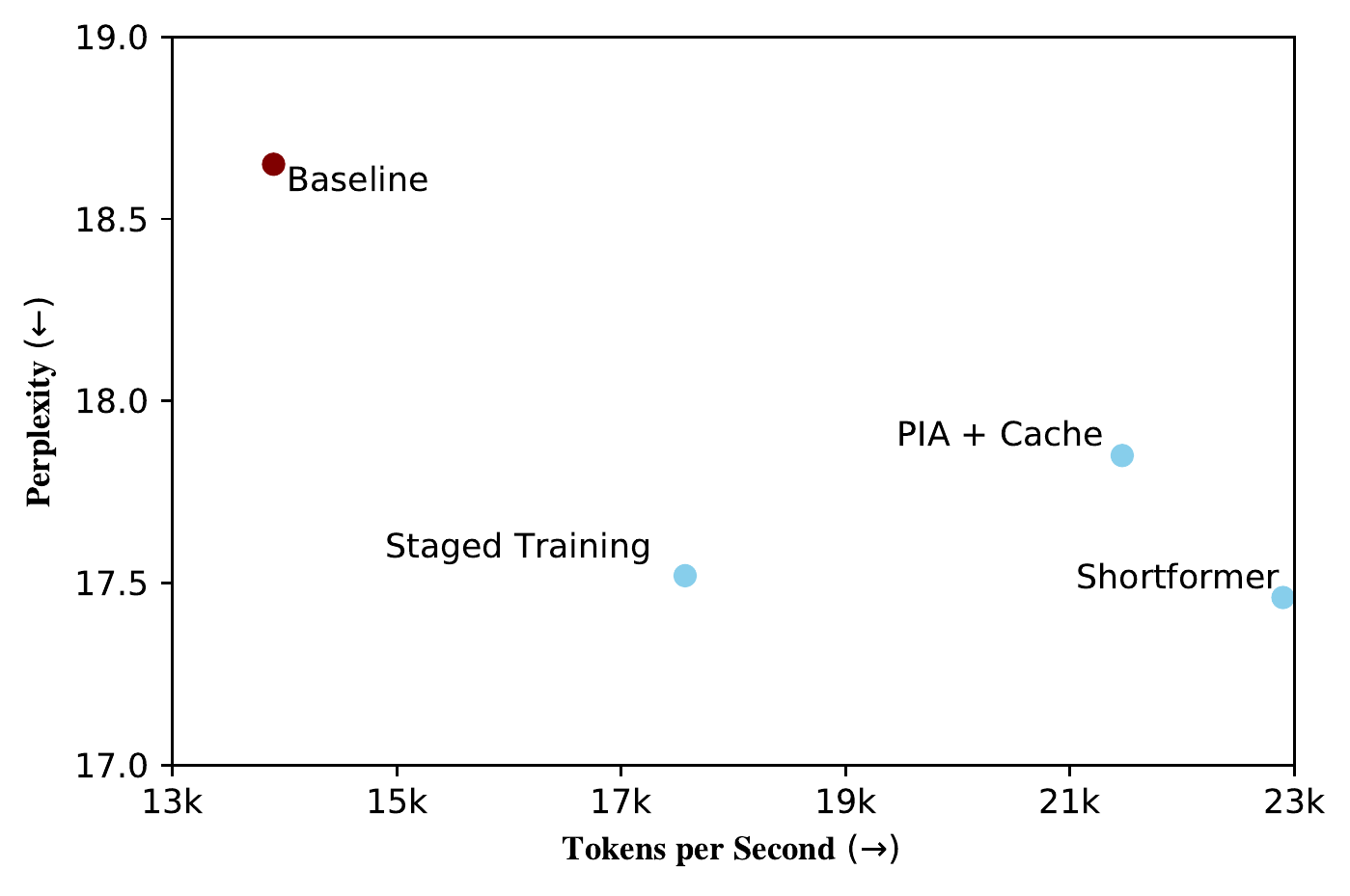}
  
 \caption{ \label{fig:all} Dev.~perplexity vs.~training speed for the baseline and our best staged training model, our best PIA and caching model, and our best combined model (Shortformer). All models are evaluated using nonoverlapping evaluation.  
 } 
\end{figure}

Figure~\ref{fig:all} compares the validation performance and training speed of the baseline to our models. 

\subsection{Memory Usage} \label{sec:mem}
\begin{table}[htp]
\centering
\small
\setlength{\tabcolsep}{5pt}

\begin{tabular}{@{}lccc@{}} \toprule
\multicolumn{2}{l}{}                                        & \multicolumn{2}{c}{Training}                             \\ \cmidrule(lr){3-4} 
      &                            & Max      & Max   \\ 
\multicolumn{2}{l}{Model} & Batch Size $\uparrow$ &Predictions  $\uparrow$      \\ \midrule
\multicolumn{2}{l}{\multirow{2}{*}{Baseline}}               & \multirow{2}{*}{2}  & \multirow{2}{*}{6,144}  \\
\multicolumn{2}{l}{}                                        &                     &                         \\ \midrule
\multirow{3}{*}{\hspace{1.35mm} Staged Training}& Stage 1   & \textbf{230}                 & \textbf{29,440}                  \\
                                 & \multirow{2}{*}{Stage 2} & \multirow{2}{*}{2}  & \multirow{2}{*}{6,144}  \\
                                 &                          &                     &                         \\ \midrule
\multicolumn{2}{l}{\multirow{2}{*}{PIA + Caching}}          & \multirow{2}{*}{26} & \multirow{2}{*}{13,312} \\
\multicolumn{2}{l}{}                                        &                     &                         \\ \midrule
\multirow{3}{*}{\hspace{1.35mm} Shortformer}     & Stage 1                  & 160 & 20,480                  \\ 
                                 & \multirow{2}{*}{Stage 2} & \multirow{2}{*}{26} & \multirow{2}{*}{13,312} \\
                                 &                          &                     &                         \\ \bottomrule
\end{tabular}
\caption{\label{tab:combined_mem} Memory usage of the baseline and our models during WikiText-103 training. For each model we show the maximal batch size that it could fit on one GPU at once during training. The max predictions column denotes the number of tokens predicted at each feedforward pass, which we calculate by multiplying batch size by number of predictions per subsequence (which is equivalent to $L$). We benchmarked all models on a V100 GPU, with 32GB of memory. Note that the second stage in the staged training model matches the performance of the baseline model, because those architectures are identical. The same is true for the second stage of the Shortformer and the PIA + Caching model.  }  
\end{table}

To understand how much memory our models and the baseline use during training, we find the largest batch size that we can load into memory for both our models and the baseline. Models that can simultaneously make more predictions are more memory efficient. 

Table~\ref{tab:combined_mem} shows the memory usage for the baseline model and our models. Since our  Shortformer model has much smaller attention matrices, during training it can make more than double the next-token predictions than the baseline can in each feedforward pass. During inference, we use a batch size of 1 throughout the paper, following~\cite{transformer-xl}, and in our experiments, the PIA + Caching model, the final staged training model and the baseline all use a similar amount of memory during nonoverlapping evaluation.
During token-by-token inference, the maximum number of predictions for the baseline model is 7, whereas our model can fit a batch size of 39 (so 39 predictions are made during token-by-token inference), making our model more than 5 times more memory efficient than the baseline. 
Using a batch size of one is a realistic benchmarking scenario: in large models such as GPT-3, a batch size of one is used during inference.

\subsection{Toronto Book Corpus} \label{sec:books}
To verify that our results transfer to other datasets, we ran our models on the Toronto Book Corpus (TBC)~\cite{zhu2015aligning}, a 700M token collection of books that has previously been used in the training corpus of BERT (along with English Wikipedia).
We use the same train/development/test split as~\cite{khandelwal20generalization} and~\cite{sandwich}, as well as their tokenization, which uses BERT’s vocabulary of 29K BPE subwords. As in ~\cite{khandelwal20generalization} and~\cite{sandwich}, since the vocabulary is much smaller than WikiText-103’s, we use a tied word embedding and softmax matrix~\cite{tying,inan2017}, instead of using the adaptive word embeddings~\cite{baevski} as in the WikiText-103 models. 

To fairly compare our models to the ones from~\cite{khandelwal20generalization} and~\cite{sandwich}, our initial set of experiments on the TBC use a maximum subsequence length of $1{,}024$ (for staged training), train for 59 epochs, and for all other hyperparameters we use the same values as the ones we used for WikiText-103 (see Experiment Setup in Section \ref{sec:background}). In this setting, the baseline achieves a perplexity of 15.38 $\pm$ 0.39 (standard deviation) on the development set. 

We do not tune the hyperparameters of our methods on the TBC, we simply use the same values as the best ones that we found on the WikiText-103 dataset. For staged training, our best model trained for $\frac{50}{205}\%$ of the epochs with $L=128$ and spent the rest of training with the same subsequence size as the baseline. 
For the TBC, we again trained the staged training model model with $L=128$ for the first $\frac{50}{205}\%$ of training, and then move on to $L=1{,}024$, to match the Sandwich Transformer~\citep{sandwich} and  kNN-LM~\citep{khandelwal20generalization} which used $1{,}024$ as the subsequence length. 

For the PIA + Caching model, we set $L=512$, as we did for our best PIA + Caching on the WikiText-103 dataset.

For the Toronto Book Corpus Shortformer, we trained for the first half of training with $L=128$ before moving on to training with $L=512$, as in our WikiText-103 models (Section~\ref{sec:combined}). 

\begin{table}[h]
\centering
\small
\setlength{\tabcolsep}{2pt}
\begin{tabular}{@{}lccccc@{}} \toprule
& \multicolumn{1}{c}{Train} & \multicolumn{4}{c}{Inference} \\  \cmidrule(lr){2-2} \cmidrule(lr){3-6}
&& &&\multicolumn{2}{c}{PPL $\downarrow$} \\ \cmidrule(lr){5-6}
Model  & Speed$\uparrow$ & Mode &  Speed$\uparrow$ & Dev.  & Test\\ \midrule
\multirow{2}{*}{Baseline} & \multirow{2}{*}{\textbf{24.0k}} & N.o. &\textbf{19.2k}& 15.38& 12.73  \\
                &               &S.W.  &9.6k&  14.75    & 11.89  \\
kNN-LM$^*$      &\textbf{24.0k} &S.W.  &-&14.20    & 10.89  \\ 
Sandwich T.     &\textbf{24.0k} &S.W.  &9.6k& -  & 10.83  \\ \midrule
PIA + Caching   &20.5k           & N.o.&15.0k   & 13.86   &  11.20  \\
\multirow{2}{*}{Staged Training }&\multirow{2}{*}{25.5k} & N.o. &\textbf{19.2k} & 13.81 & 11.18 \\
& &S.W.  &9.6k & \textbf{13.13}   &  \textbf{10.72} \\
Shortformer     &21.3k          & N.o.  &15.5k& 13.40    & 10.88  \\ \bottomrule

\end{tabular}
\caption{\label{tab:combined_test_books} Comparison of our best models to other strong LMs trained on the Toronto Book Corpus (TBC). Following~\citet{khandelwal20generalization} and~\citet{sandwich}, for the baseline and our staged training model, we set  $L=1{,}024$ and $S=512$ when using sliding window (S.W.) evaluation in the TBC dataset. All models have 261M parameters. $^{*}$kNN-LM requires a 400GB datastore. }
\end{table}

Table~\ref{tab:combined_test_books} shows that staged training and the Shortformer improve over the baseline by a wide margin and match the results of the Sandwich Transformer and the kNN-LM. As noted in Section~\ref{sec:combined}, those contributions are orthogonal to ours, and combining them might yield further gains. 
Since in Table~\ref{tab:combined_test_books} the final stage of the staged training model (and the baseline) both have $L = 1{,}024$, Shortformer lacks a speed advantage in this scenario. 

Table~\ref{tab:combined_test_books_3k} shows results for our staged training model trained with a final stage subsequence length of $3{,}072$ tokens, as in our WikiText-103 experiments in Section~\ref{sec:ss}. This model trains faster than the $L=3{,}072$ baseline and also achieves much better perplexity scores (the baseline in this setting achieves a perplexity of 14.52 $\pm$ 0.15 (standard deviation) on the development set). In addition, note that the Shortformer model from Table~\ref{tab:combined_test_books} achieves better perplexity than even the baseline with $L=3{,}072$, although Shortformer is much faster to train and uses much smaller attention matrices during inference (of size $512\cdot 1024$; the baseline has attention matrices of size $3{,}072 \cdot 3{,}072 $, as in Section~\ref{sec:combined}).

\begin{table}[h]
\centering
\small
\setlength{\tabcolsep}{2pt}
\begin{tabular}{@{}lccccc@{}} \toprule
& \multicolumn{1}{c}{Train} & \multicolumn{4}{c}{Inference} \\  \cmidrule(lr){2-2} \cmidrule(lr){3-6}
&&& &\multicolumn{2}{c}{PPL $\downarrow$} \\ \cmidrule(lr){5-6}
Model  & Speed$\uparrow$ & Mode & Speed$\uparrow$ & Dev.  & Test\\ \midrule
 Baseline & \multirow{2}{*}{14.2 } & N.o. &\textbf{15.1}&14.52  & 11.69  \\
($L=3{,}072$)       &  &S.W.&2.5k &14.14    &11.43   \\ \midrule

Staged Training  & \multirow{2}{*}{\textbf{18.1}}   & N.o.     &\textbf{15.1}& 13.19  &  10.76 \\ 
($L=3{,}072$)   &   & S.W.  &  2.5k & \textbf{12.80}  &  \textbf{10.48} \\\bottomrule

\end{tabular}
\caption{\label{tab:combined_test_books_3k} Comparison of the staged training model to the baseline, when the subsequence length $L$ is set to $3{,}072$. In this table, $S=512$. }
\end{table}

\end{document}